\documentclass[lettersize,journal]{IEEEtran}
\usepackage{amsmath,amsfonts}
\usepackage{algorithmic}
\usepackage{algorithm}
\usepackage{array}
\usepackage[caption=false,font=normalsize,labelfont=sf,textfont=sf]{subfig}
\usepackage{textcomp}
\usepackage{stfloats}
\usepackage{url}
\usepackage{verbatim}
\usepackage{graphicx}
\usepackage{cite}
\newcommand{\argmin}{\operatornamewithlimits{arg\ min}}
\usepackage{siunitx}

\begin{document}

\title{OpenStreetMap-based Autonomous Navigation\\ With LiDAR Naive-Valley-Path Obstacle Avoidance}

\author{Miguel Ángel Muñoz-Bañón, Edison Velasco-Sánchez, Francisco A. Candelas,\\ and Fernando Torres~\IEEEmembership{Senior member,~IEEE,} 
% <-this % stops a space
\thanks{This work has been supported by the Spanish Goverment through the FPI grant PRE2019-088069 and the research project RTI2018-094279-B-100, as well as by the regional Valencian Community Goverment and the European Regional Development Fund (ERDF) through the grant ACIF/2019/088.}
% <-this % stops a space
\thanks{The authors are with the Group of Automation, Robotics and Computer Vision (AUROVA), University of Alicante, San Vicente del Raspeig S/N, Alicante, Spain. {\tt\small miguelangel.munoz@ua.es}}
\thanks{Digital Object Identifier (DOI): 10.1109/TITS.2022.3208829}
}

% The paper headers
\markboth{IEEE TRANSACTIONS ON INTELLIGENT TRANSPORTATION SYSTEMS, VOL. 23, NO. 12, DEC. 2022}%
{Shell \MakeLowercase{\textit{et al.}}: OpenStreetMap-based Autonomous Navigation\\ With LiDAR Naive-Valley-Path Obstacle Avoidance}

\maketitle

\begin{abstract}
OpenStreetMaps (OSM) is currently studied as the environment representation for autonomous navigation. It provides advantages such as global consistency, a heavy-less map construction process, and a wide variety of road information publicly available. However, the location of this information is usually not very accurate locally. 

In this paper, we present a complete autonomous navigation pipeline using OSM information as environment representation for global planning. To avoid the flaw of local low-accuracy, we offer the novel LiDAR-based Naive-Valley-Path (NVP) method that exploits the concept of "valley" areas to infer the local path always furthest from obstacles. This behavior allows navigation always through the center of trafficable areas following the road's shape independently of OSM error. Furthermore, NVP is a naive method that is highly sample-time-efficient. This time efficiency also enables obstacle avoidance, even for dynamic objects.

We demonstrate the system's robustness in our research platform BLUE, driving autonomously across the University of Alicante Scientific Park for more than 20 km with 0.24 meters of average error against the road's center with a 19.8 ms of average sample time. Our vehicle avoids static obstacles in the road and even dynamic ones, such as vehicles and pedestrians.
\end{abstract}

\begin{IEEEkeywords}
Autonomous navigation, Unmanned ground vehicle, Open street maps, Path planning, Obstacle avoidance, Lidar point cloud.
\end{IEEEkeywords}

%% main text
%%%%%%%%%%%%%%%%%%%%%%%%%%%%%%%%%%%%%%%%%%%%%%%%%%%%%%%%%%%%%%%%%%%%%%%%%%%%%%%%

\begin{figure}[t]
\centering
\includegraphics[width=200pt]{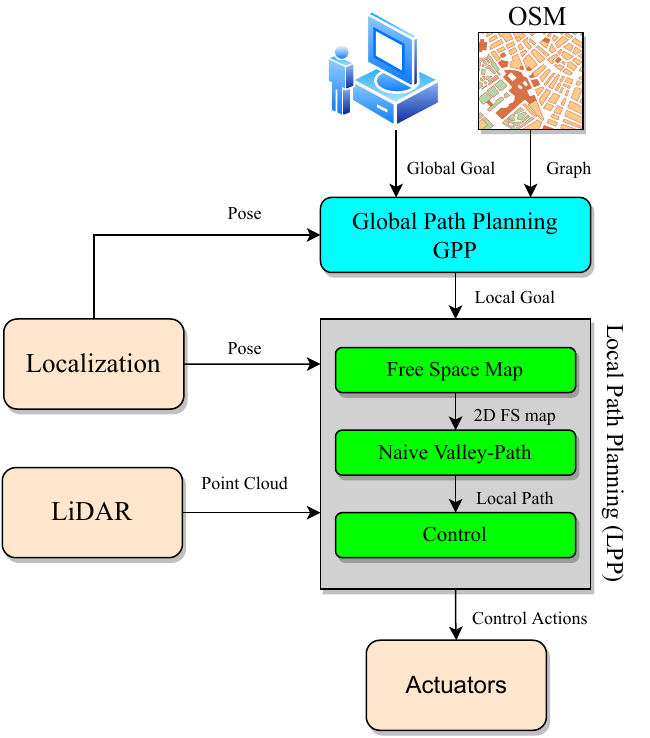}
\caption{The complete autonomous navigation pipeline, where path planning approach is organized hierarchically. First, the GPP gets information about global localization from the robot and global information about the environment from OSM to plan a global path. Then, the LPP module, using the presented Naive-Valley-Path (NVP) method, recalculates the local path to obtain the optimal way to follow.}
\label{fig:overview}
\end{figure}

\section{Introduction}
\label{}

\IEEEPARstart{A}{utonomous} navigation aims to reach a specific target without human intervention. This task requires knowledge about the environment to infer the vehicle's pose \cite{thrun2001robust} and to plan the path for goal-reaching \cite{gonzalez2015review}. For past decades, the problem of generating a model of the environment has been widely studied. \textit{Simultaneous Localization And Mapping} (SLAM) \cite{cadena2016past} is one of the most researched approaches in the literature. The SLAM aims to incrementally generate a map through an unknown environment, using the map for localization simultaneously. However, the SLAM maps usually provide high local accuracy but no global consistency.  Such maps often show accumulated drifts and scale errors. GNSS sensors don't avoid these limitations due to multipath, shadowing, and atmospheric drift issues \cite{lesouple2018multipath}. Such issues produce deviations, in some cases, in the order of meters. Also, the errors usually have different magnitudes depending on the position and even the hour of the day, producing shifted and smoothly distorted maps. A significant undesirable effect of that behavior is that the maps created in different sessions usually don't coincide in their joint parts. Then, it is hard to merge them in a common coordinates frame. In that case, the global path planning becomes unrealizable due to the maps being unconnected. Furthermore, SLAM-based systems require a previous exploration of the work area that produces an increment of the application complexity \cite{toriz2020mejora}, Especially when the area size increases and when exploring new places becomes necessary.

As a SLAM alternative, OpenStreetMaps (OSM) has been studied in the last years. OSM is a knowledge collective that provides user-generated street maps publicly available \cite{haklay2008openstreetmap}. Its use yields different advantages, such as global consistency, the simplicity of the map construction process that lightens the application's setup, and the amount of publicly available data. Given these benefits, there has been a wide variety of OpenStreetMaps-based autonomous navigation works, such as \cite{alonso2012accurate,floros2013openstreetslam,naik2019semantic}, where the authors use OSM information for localization and even, in the case of \cite{naik2019semantic}, for mapping. In \cite{artunedo2017smooth,li2021openstreetmap,chen2016crowddeliver}, the authors use road networks from OSM for Global Path Planning (GPP). Due to OSM graph-like representation, the most common approach used in that context for GPP is graph search algorithm \cite{kularatne2017optimal,chen1998sandros,mannarini2019graph} and its variants, such as the Dijkstra algorithm \cite{broumi2016applying}, A* algorithm \cite{duchovn2014path}, DFS algorithm \cite{guo2013revising}, and BFS algorithm \cite{yu2013planning}. However, while OSM is well-fitting to the GPP problem due to its high global consistency, it is, in fact, locally inaccurate and usually entails a shift in the road representation. This behavior can produce deviations in the local paths inferred by the Local Path Planning  (LPP) module.

In LPP approaches such as sample-based methods \cite{gammell2021asymptotically,lim2018hierarchical},  the local environment is represented as a cost map sampled, usually as a grid map, built from LiDAR or cameras data. Then, the local path that minimizes the cost in the map is searched using algorithms such as A* \cite{li2021openstreetmap} or RRT* \cite{chi2018risk}. The cost maps are generally very informative about the obstacles and the limits of the map but not at the center of trafficable areas. Then, the non-center-informative LPP and the previously mentioned local inaccuracy of OSM-based GPP can produce lateral deviations in the trajectory. Furthermore, these lateral deviations can also occur in the case of sparse road network representation, where the vehicle would not follow the trajectory with the same shape as the road.

In some OSM-based works, the authors focus on improving localization systems by fusion techniques \cite{zaman2013integrated} or by sensor-based perception \cite{ort2018autonomous}. Nevertheless, they don't consider the possible OSM inaccuracy, and the local trajectory can suffer deviations. In \cite{suger2017global}, the authors correct the OSM global trajectory by fitting it to the previously segmented road using a 3D-LiDAR sensor. However, the correction depends strongly on the road segmentation that could not be robust enough. In a recent work \cite{li2021openstreetmap}, the authors use a similar approach by correcting the OSM path using a cost map built, in this case, using combined camera and LiDAR information. As LPP, to follow the corrected OSM path, the authors implement A* using the cost map. In the previously-mentioned approaches \cite{suger2017global,li2021openstreetmap}, each node from the OSM road network is corrected when the vehicle follows them. But due to possible localization errors, this correction can not be permanent, which entails that the methods should relocate the path in each autonomous navigation session. Given this assumption, we consider more elegant to maintain OSM information constant for GPP and avoid the problem of deviations in the LPP module. 

In this paper, we address the local inaccuracies that produce deviations problems in OpenStreetMap-based autonomous navigation systems by presenting the novel LiDAR-based Naive-Valley-Path (NVP), an LPP approach. It is worth noting that the NVP is not presented as a contribution to the general LPP problems but as a contribution applied to improve OSM-based systems, and we will evaluate it in that context. This method uses a potentials local environment representation that exploits the concept of "valley" areas, which have lower gradient values. Such areas always follow trafficable road shapes, avoiding the common deviations in the OSM-based applications. This work is developed in the context of a complete OpenStreetMap-based autonomous navigation pipeline using OSM road network as environment representation for GPP\footnote{This work is in the context of a real application for a project that addresses the problem of garbage "pick and place" in the University of Alicante campus using Unmanned Ground Vehicles (UGV). For this reason, we assume an unstructured outdoor environment for our application, where the localization is GPS-IMU based, in a global frame coordinates system.}. The presented method is a "naive" version very efficient in computational time terms, in contrast with the commons sample-based methods \cite{li2021openstreetmap}. Due to this time efficiency, we can achieve real-time obstacle avoidance, even for dynamic obstacles.

To summarize, the main contributions are the following:
\begin{itemize}

    \item A novel real-time method so-called  Naive-Valley-Path (NVP). That LPP is developed to avoid local inaccuracies of OSM-based autonomous navigation systems. This method infers a naive cost map represented as concentric circles around the robot to obtain the optimal local path using points in "valley" areas. It provides two main advantages: navigation always following the center of the trafficable regions, avoiding the common deviation problems in OSM-based applications, and low execution time.
    \item A complete outdoor autonomous navigation system for unstructured environments, based on GPS-IMU fusion localization, OSM for GPP, and a sampled-based LPP for road center correction and obstacle avoidance using LiDAR measurements.
    \item Test and comparison with other state-of-the-art OSM-based autonomous navigation: \textit{Li et al.} \cite{li2021openstreetmap}. To perform that OSM-based implementations, we use our own developed research platform BLUE \cite{del2020deeper} and our navigation framework \cite{munoz2019framework}.
\end{itemize}

The rest of the paper is organized as follows: In Section \ref{sec:proposed_approach}, we present an overview of the complete autonomous navigation pipeline proposed. Then, local Path Planning and Global Path Planning modules are described in Sections \ref{sec:global_planning} and \ref{sec:local_planning}, respectively. Next, in Seccion \ref{sec:experiments}, we show the experimental results obtained using our own real robot BLUE. Finally, in Section \ref{sec:conclusions}, we present the main conclusions obtained from this work and possible future works.

%%%%%%%%%%%%%%%%%%%%%%%%%%%%%%%%%%%%%%%%%%%%%%%%%%%%%%%%%%%%%%%%%%%%%%%%%%%%%%%%

\section{Proposed Approach Architecture}
\label{sec:proposed_approach}

In Fig. \ref{fig:overview}, we show the proposed approach, which is divided into different modules. The path planning is organized hierarchically. At the top-level, we compute the GPP that receives a graph obtained from the online application OSM, the final global goal provided by the user, and the pose obtained from the localization module. In this work, we assume that we have a georeferenced global localization. This is required to use the environment representation directly from OSM, which provides geolocalized information about roads and intersections. The GPP infers the best path through the graph using the A* algorithm, and it gives as output a local goal, which is the nearest node of the calculated path. 

This goal is an input for the LPP module. Moreover, the LPP receives as inputs the vehicle's localization and 3D LiDAR scans from the sensor. We divided the  LPP  hierarchically into three layers. In the ”Free-space” calculation layer, we remove ground points and the points above the upper part of the vehicle. Then, we consider the rest of the points as obstacles that we use as borders of a free-space representation 2D map. In the Naive-Valley-Path NVP layer, we use a naive cost map representation in the free-space map to infer an optimal path exploiting the concept of "valley" areas. Finally, we evaluate the possible actions, among ones without collision risk, that minimize the error between the vehicle pose and the local path.

In Section \ref{sec:global_planning} and Section \ref{sec:local_planning}, we explain in more detail each module of the proposed approach.

%%%%%%%%%%%%%%%%%%%%%%%%%%%%%%%%%%%%%%%%%%%%%%%%%%%%%%%%%%%%%%%%%%%%%%%%%%%%%%%%

\begin{figure}[t]
\centering
\includegraphics[width=200pt]{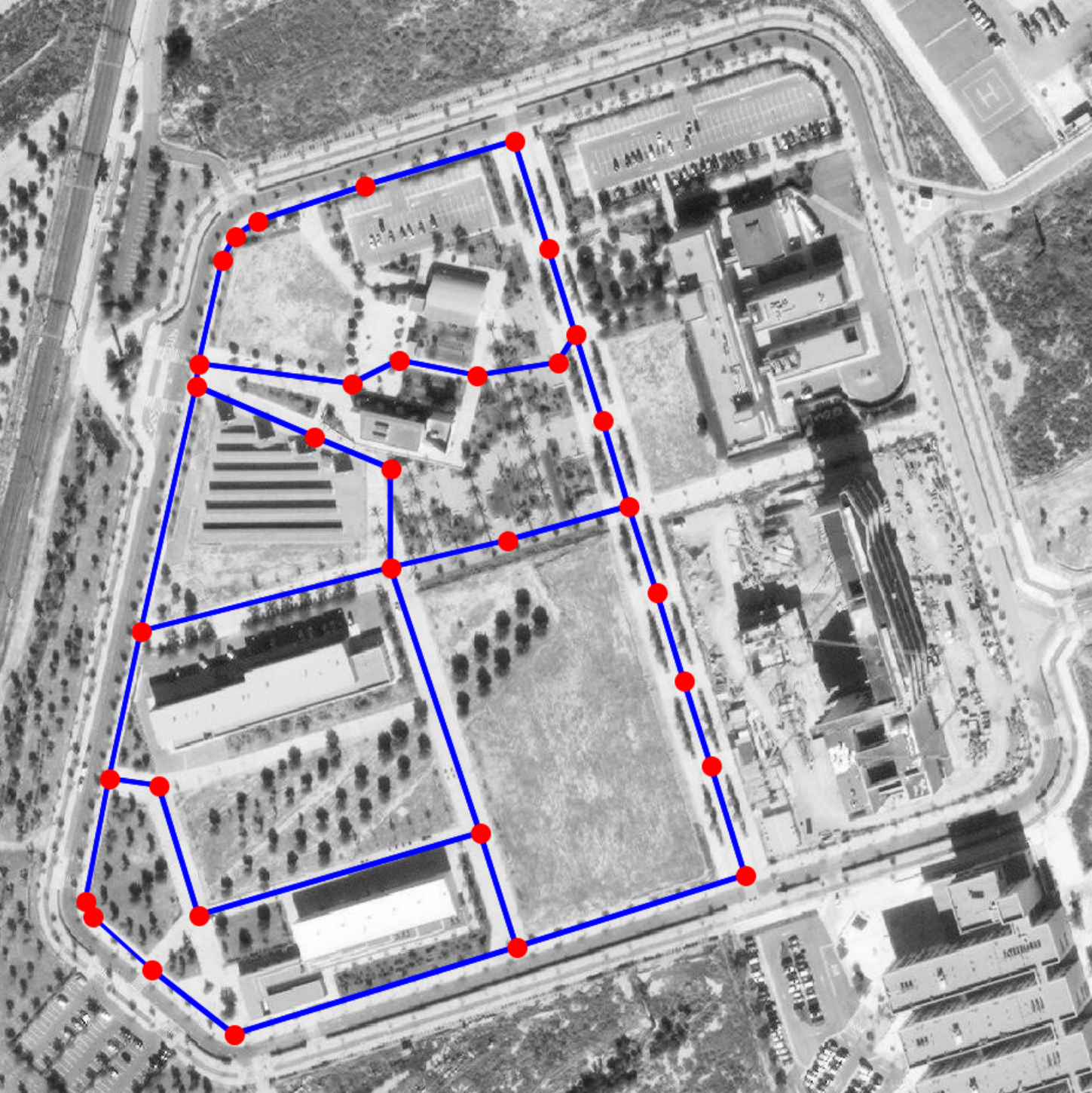}
\caption{Example of road network (graph) used in the GPP module, in this case, extracted from the Scientific Park area in the University of Alicante. The red points describe the georeferenced position of nodes, while the blue lines represent the links that indicate a trafficable connection between nodes. We can extract the graph directly from current OSM data or create it manually in the JOSM application.}
\label{fig:graph}
\end{figure}

\section{Global Path Planning}
\label{sec:global_planning}

At the GPP level, we represent the environment as a road network described as a graph $\mathcal{G} = (\mathcal{W}, \mathcal{A})$, where $\mathcal{W}$ is a set of nodes, and $\mathcal{A}$ is a set of links defined over nodes. In such representation, each node $\mathbf{w}_i \in \mathcal{W}$ is a georeferenced point in a trafficable area. Each node contains information about latitude, longitude, and unique identification. Each link $\mathbf{a}_{ij} \in \mathcal{A}$ indicates that two points are connected through a passable road. In Fig. \ref{fig:graph}, we show an example of a graph built from OSM information and plotted over an aerial image. We can obtain the graph representation by downloading directly from OSM. However, in this work, following the roads' shape falls on the LPP module. For this reason, if the roads described in OSM are dense, we can subsample them to give a more sparse representation. Additionally, we can add nodes manually to areas that we know are trafficable using the software of OSM called JOSM.

We can define $\mathbf{w}_{start} \in \mathcal{W}$ as the nearest node to the vehicle's pose $\mathbf{p}_{vehicle}$. And $\mathbf{w}_{end} \in \mathcal{W}$ as the node closest to the global goal $\mathbf{g}_{g}$ provided by the user. Then, given a graph $\mathcal{G}$, we find the best global path $\mathbf{P}^g$ between $\mathbf{w}_{start}$ and $\mathbf{w}_{end}$ using the A* algorithm. Then, we describe the global path as a set of waypoints $\mathbf{P}^g = (\mathbf{w}_1^g, \mathbf{w}_2^g, ..., \mathbf{w}_n^g)$. Once the path $\mathbf{P}^g$ has been obtained, we store the waypoints in a buffer. Finally, we send each waypoint sequentially as a local goal $\mathbf{g}_l = \mathbf{w}_i^g$ to the LPP module as they are being reached. We consider a local goal reached when the Mahalanobis Distance (MD) \cite{de2000mahalanobis} is less than a certain configurable threshold. We measure the MD between the graph node (local goal $\mathbf{g}_l$) and the localization distribution $\mathbf{X} \sim N(\mathbf{x}, \mathbf{\Sigma})$, where $\mathbf{x}$ is the vehicle's pose, and $\mathbf{\Sigma}$ is the covariance. Using this probabilistic distance, we can evaluate if a goal is reached depending on the covariance $\mathbf{\Sigma}$ in the localization system. In this way, we can prevent severe deviation from achieving goals in considerable noisy localization.

%%%%%%%%%%%%%%%%%%%%%%%%%%%%%%%%%%%%%%%%%%%%%%%%%%%%%%%%%%%%%%%%%%%%%%%%%%%%%%%%

\begin{figure*}[t!]
\centering
\includegraphics[width=450pt]{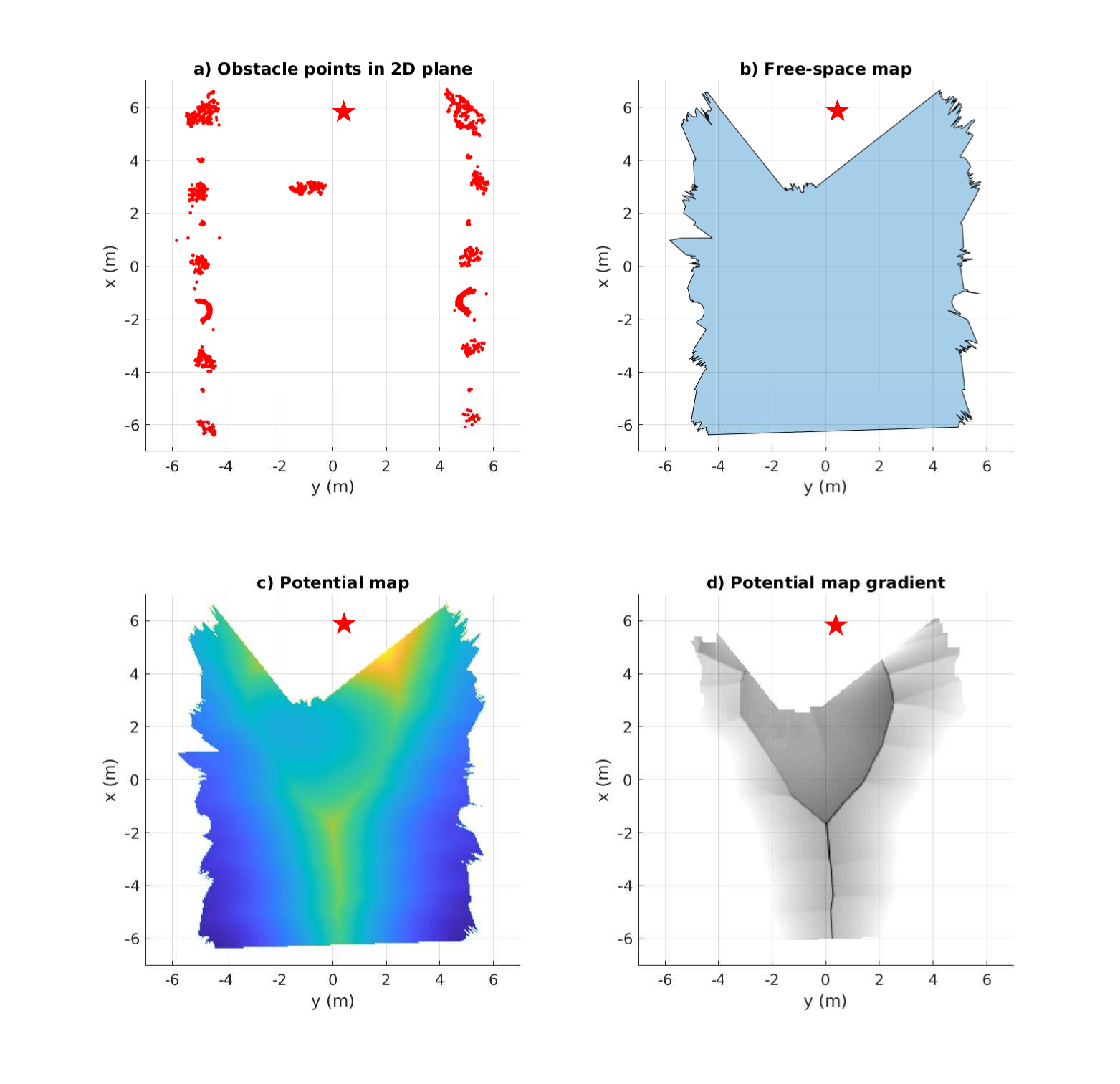}
\caption{Different top-view 2D representations of LiDAR information for Valley-Path calculation given a target goal represented as a red star: a) Projection in a top-view 2D plane of obstacles point cloud $\mathcal{P}^o$. b) Free-space map, where the blue area represents the space free of obstacles. c) Cost map defined in (2). d) Inverted representation of the gradient magnitude of c), which shows clear possible paths in the valley areas.}
\label{fig:maps}
\end{figure*}

\section{Local Path Planning}
\label{sec:local_planning}

In this section, we describe the Local Path Planning module, which is divided hierarchically into three levels. This module aims to infer the final control actions to send to the vehicle's actuators. Also, in this process, we use the LiDAR sensor to follow the center of the road and avoid possible obstacles in the local goal-reaching process.

\subsection{Obstacles and Free-space Calculation}

We consider as obstacles the points in a LiDAR point cloud that are not part of the surface on which the vehicle circulates, i.e., the ones that are representing objects above the ground, or even points under the ground, such as descending steps. Besides, we consider obstacles only the points that are under the upper part of the vehicle, which is the collision risk fringe.

Given this definition, we need to detect the ground points to consider obstacles to the rest. In large part of the scenarios, we observed the ground surface is usually flat. Hence, we make a plane assumption using the LiDAR point cloud $\mathcal{P}$ for a nonlinear Least-Squares optimization (based on \cite{wirges2018evidential}) to find the optimal ground plane parameters (1) that minimize the accumulated point-to-plane error for all points $\mathbf{p} \in \mathcal{P}$:

$$
\mathbf{pl}^* = \argmin_{\mathbf{pl}} \sum_{\mathbf{p} \in  \mathcal{P}} \rho \left(||\mathbf{e}(\mathbf{pl}, \mathbf{p})||^2 \right) \eqno{(1)}
$$

Where $\mathbf{e}(\mathbf{pl}, \mathbf{p})$ denotes the distance vector between $\mathbf{p}$ and its plane projection point. The loss function $\rho$ is chosen to be the Cauchy loss with a small scale to be robust against outliers. We then remove all points from the point set with a distance below a threshold to the $\mathbf{pl}^*$ plane. To consider possible slope changes in the terrain, we apply this threshold proportionally to the distance to the sensor center. Finally, we displace the plane in the $z$ axis to the upper part of the vehicle, and we then remove the points over this second plane. The points that remain in $\mathcal{P}$ after this process are what we consider obstacles in a new point cloud $\mathcal{P}^o$. In Fig. \ref{fig:maps} a), we show an example of $\mathcal{P}^o$ projected in a top-view 2D plane.

Starting from $\mathcal{P}^o$, we build a 2D representation of the free space. We define free space as the area inside a polygon where the contours are obstacle points. We implement the polygon calculation in a two-stage process. In \textbf{stage 1}, each Cartesian point $(x, y, z) \in \mathcal{P}^o$ is transformed into spherical coordinates $(\phi, \theta, \rho)$. Then, we use $\phi$ and $\theta$ as an index to build a front-view image $\mathbf{I}_{FV} \in \mathbb{R}^{H \times W}$, where $H$ and $W$ depend on the LiDAR resolution and the range limits of $\phi$ and $\theta$ respectively. The value in each cell of $\mathbf{I}_{FV}$ is the range value $\rho$. The cells with no point information have an empty value. In Section IV-A of \cite{vaquero2020dual}, the authors derive with more detail this representation. In \textbf{stage 2}, we sweep the columns in $\mathbf{I}_{FV}$. If the column contains non-empty points, we choose the one with the lowest $\rho$ value, named $\rho'$ from now on. Then, for each selected point for each non-empty column, we transform $(\phi, \theta, \rho)'$ into a top-view 2D Cartesian representation $(x,y)$. The points selected are the ones that form the free-space polygon. In Fig. \ref{fig:maps} b), we show an example of a free-space map.

\begin{figure}[t!]
\centering
\includegraphics[width=220pt]{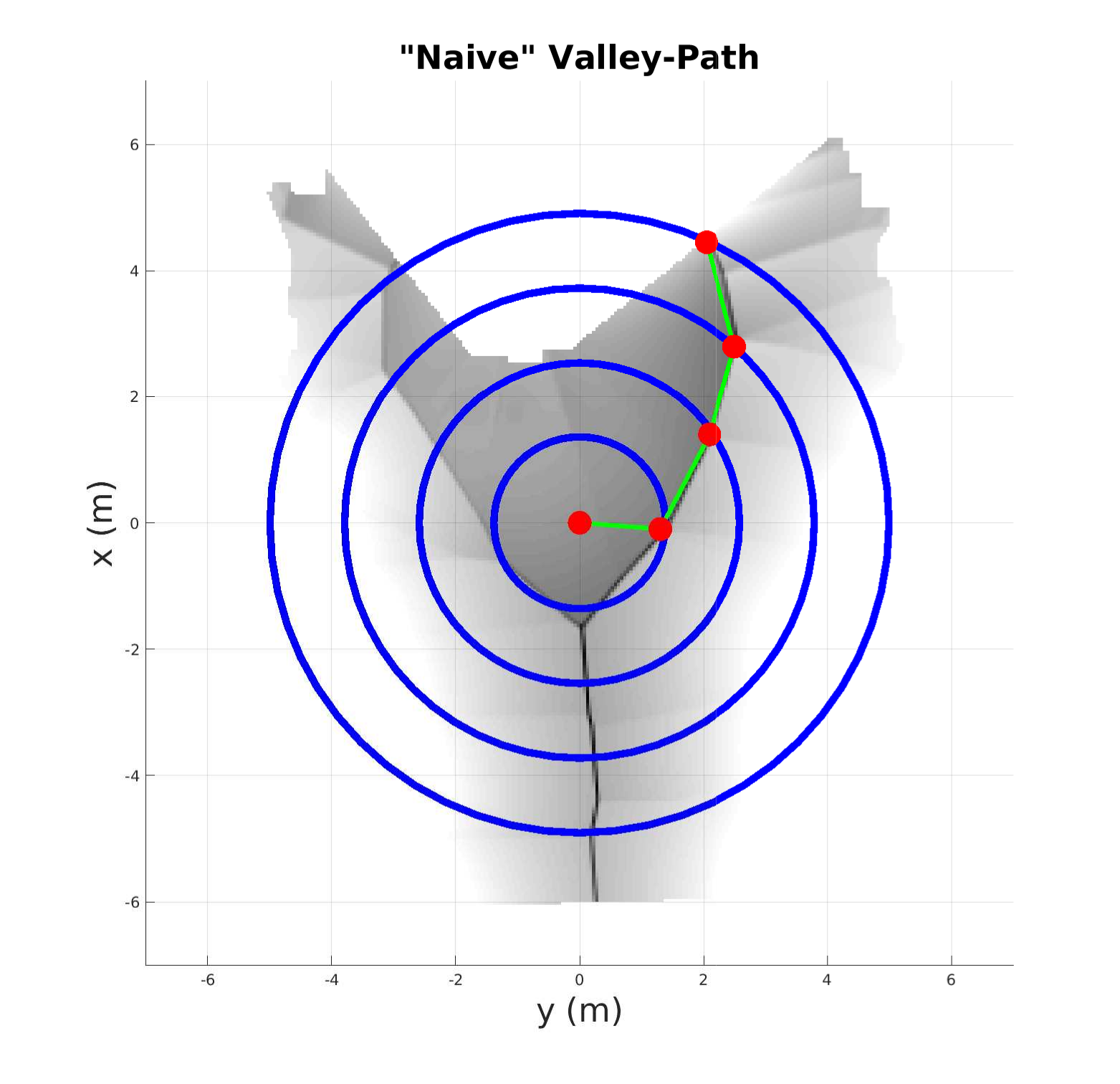}
\caption{The Naive-Valley-Path (NVP) Calculation. The red points are the ones that form the local path $\mathbf{P}^l = \left(\mathbf{p}^{c_0}_{nn}, \mathbf{p}^{c_1}_{nn}, ..., \mathbf{p}^{c_N}_{nn} \right)$. The green connections describe the \textbf{naive assumption} and define the angle of the points. For the sake of clarity, this representation shows the NVP superpose with non-naive representation.}
\label{fig:naive_path}
\end{figure}

\subsection{Valley-Path Calculation}

Once we know the free space around the vehicle, we need to determine a path in that space to reach the local goal. For this, we use a cost representation of free space based on potentials. We consider the local goal as an attractor $\mathbf{p}_{a} = \mathbf{g}_l$ and the nearest obstacle point as a repulsor $\mathbf{p}_{r}$. Then, we can sample the space around the polygon as a grid defined as a matrix $\mathbf{F}$, where we represent each cell as $f_{ij} \in \mathbf{F}$. For each $f_{ij}$ inside the free space polygon, we obtain the cost as follows:

$$
f_{ij} = \frac{w_r}{\left\lVert \mathbf{p}_{ij} - \mathbf{p}_{r} \right\rVert^{\gamma_r}} - \frac{w_a}{\left\lVert \mathbf{p}_{ij} - \mathbf{p}_{a} \right\rVert^{\gamma_a}} \eqno{(2)}
$$

Where $w_a$ and $w_r$ are the weight of the attractive potential and the repulsive potential, respectively. And where $\gamma_a$ and $\gamma_r$ are the parameters to control the decay of the potential with respect to the distance. In Fig. \ref{fig:maps} c), we show an example of our cost map representation $\mathbf{F}$ for single LiDAR scan. We consider that to follow the shape of the road; we always want to navigate through the areas furthest from obstacles. Then, focusing on yellow areas in Fig. \ref{fig:maps} c), we can observe that these areas are ones around local minimal. We name these areas as "valleys" from now on. Following that valleys, we ensure that the local path can correct lateral deviations derived from OSM-based GPP, which is the main advantage of presented NVP.

To segment valleys given the cost map $\mathbf{F}$, we first obtain the gradient:

$$
\nabla \mathbf{F} = \left[ \frac{\partial\mathbf{F}}{\partial x}, \frac{\partial\mathbf{F}}{\partial y} \right] \eqno{(3)}
$$

Next, we can also obtain the magnitude of gradient as follows:

$$
\left| \nabla \mathbf{F} \right| = \sqrt{\left( \frac{\partial\mathbf{F}}{\partial x} \right)^2 + \left( \frac{\partial\mathbf{F}}{\partial y} \right)^2} \eqno{(4)}
$$

Then, to segment the valleys in the map, we assume that the values of the magnitude of the gradient at point $\left|\nabla f_{ij}\right| \in \left| \nabla \mathbf{F} \right|$ close to zero can be considered a point in a valley. Hence, each point that satisfies (5) is labeled as a valley. 

$$
\left|\nabla f_{ij}\right| < \xi  \eqno{(5)}
$$

Where $\xi$ is a configurable threshold. Fig. \ref{fig:maps} d) shows an example of a segmented valley. In the example, the darkest points mark the lowest magnitude of gradient points. We can see that this representation define clearly the shape of the road and possible Valley-Paths that can correct local inaccuracies of the OSM-based GPP\footnote{\textbf{Differences between Valley-Path and the classical Potential Fields method}: In PF, the potential gradient is used to infer the control actions directly. This approach usually suffers due to local minimums present in the potentials representation. In contrast, we do not infer the control actions from the potential field; we infer the local path (aka Valley-Path). Hence, the local minimums in the representation (cost map) are not an inconvenience. Conversely, the local minimums are parts we want to cross on the local path because they are furthest from obstacles.}.

To infer a Valley-Path between the robot pose and the global minimum in the map, we could use the gradient cost map for an optimization process such as A* or Dijkstra \cite{ardakani2015decremental} among others \cite{ortiz2005mathematical}. However, this approach has the flaw of being excessively time-consuming in computational terms. For this reason, we develop a "naive" version of this Valley-Path calculation that we explain in Section \ref{sec:naive_path}.

\subsection{Naive Version of the Valley-Path Calculation}
\label{sec:naive_path}

Given a circle around the sensor pose, for each polar coordinate $(r, \varphi_i)$ we can derive a 1D cost function $\mathbf{f}$, where for each $f_i \in \mathbf{f}$ we apply the expression (6), which is the 1D version of (2). 

$$
f_{i} = \frac{w_r}{\left\lVert \mathbf{p}_{i} - \mathbf{p}_{r} \right\rVert^{\gamma_r}} - \frac{w_a}{\left\lVert \mathbf{p}_{i} - \mathbf{p}_{a} \right\rVert^{\gamma_a}} \eqno{(6)}
$$

Where $\mathbf{p}_{i} = polarToCartesian(r, \varphi_i)$. If we represent this 1-dimensional signal as a magnitude of the gradient, we can label as valley points the ones that satisfy $\nabla f_{i} < \xi$. We can do the same process in inner concentric circles. In this way, we can make the following \textbf{naive assumption}: given a valley point $\mathbf{p}^{c_1}_i$ in a circle $c_1$, and given the nearest valley point $\mathbf{p}^{c_2}_{nn}$ in an inner circle $c_2$, the line that connects $\mathbf{p}^{c_1}_i$ with $\mathbf{p}^{c_2}_{nn}$ is considered part of a path in the same valley. The subscript $nn$ means the index of the nearest valley point.

Under this assumption, given a set of $N$ circles $\left(c_0, c_1, ..., c_N \right)$, we can define the local Naive-Valley-Path (NVP) as a set of join waypoints $\mathbf{P}^l = \left(\mathbf{p}^{c_0}_{nn}, \mathbf{p}^{c_1}_{nn}, ..., \mathbf{p}^{c_N}_{nn} \right)$ in which each element is connected with the nearest previous one. The first waypoint $\mathbf{p}^{c_0}_{nn}$ is the valley point in the external circle nearest to the local goal $\mathbf{p}_a$. As $N$ increases, the time consumption also increases, and the result can converge to the non-naive version of the Valley-Path calculation. The process described in this section is executed each time a LiDAR scan is received, i.e., the local path is recalculated in each iteration. In Fig. \ref{fig:naive_path} we show an example of NVP and the circles used for the inference. To more clearly show how NVP follows the valley areas, in Fig. \ref{fig:naive_path} we superpose it to the original non-naive representation.

\begin{figure}[t]
\centering
\includegraphics[width=220pt]{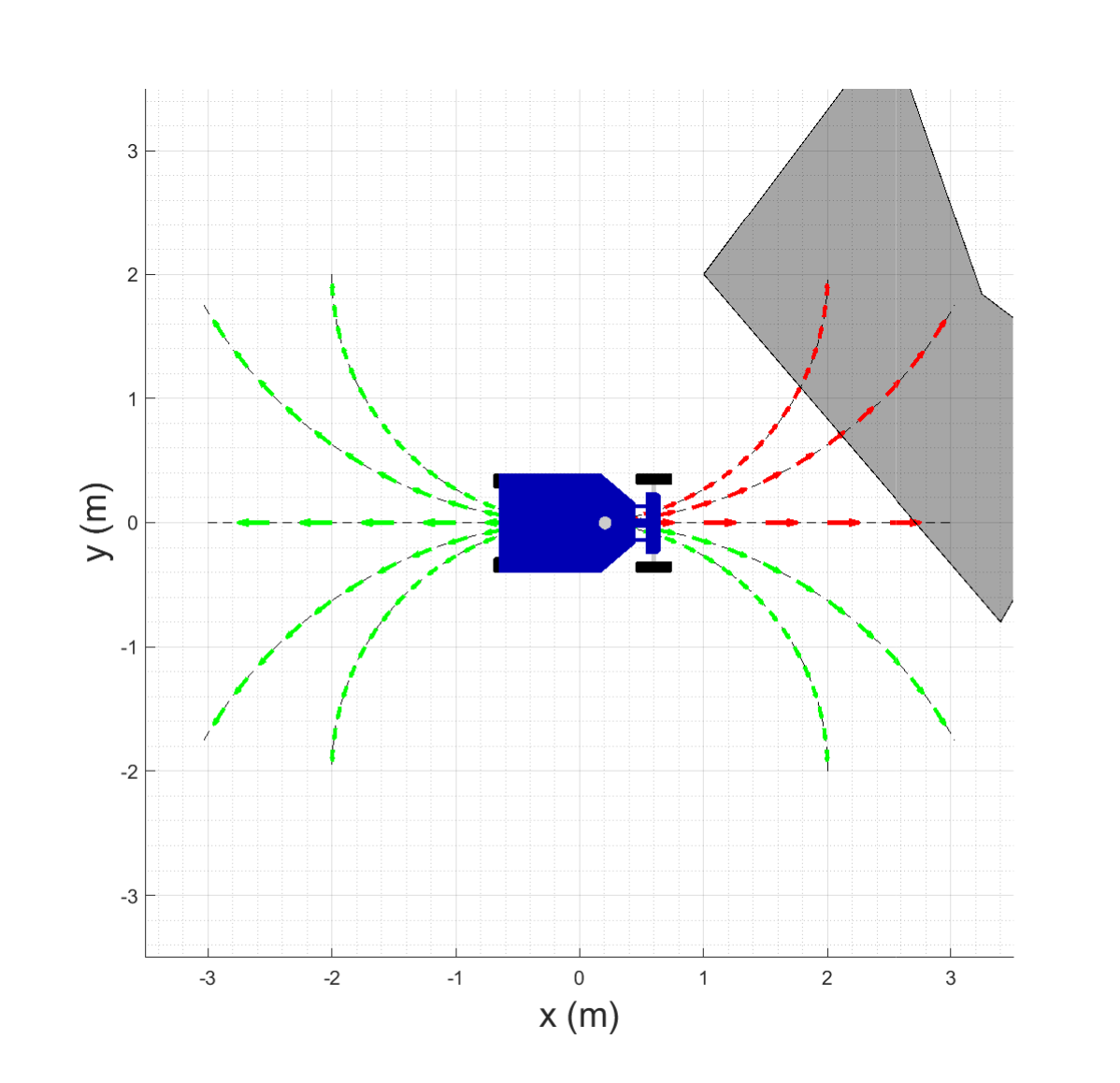}
\caption{The arcs represent the possible trajectories of control actions $\mathbf{u}_i$. The red ones are the collision-risk trajectories, and their corresponding control actions are discarded. In contrast, the green ones are collision-free.}
\label{fig:control_action}
\end{figure}

\subsection{Control Actions Calculation}

The local path calculation doesn't consider the vehicle's kinematic, then to follow that path, we implement in the lower-level layer the controller that depends on the vehicle's kinematic. 

Given a local path, we obtain the control actions $\mathbf{u} = (v,  \alpha)$ that are the output of the LPP module. $v$ is the linear absolute velocity of the vehicle, and $\alpha$ is the steering position. For this, we compute a prediction of trajectories for any possible $\alpha$ at the front and rear directions (Fig. \ref{fig:control_action}). Given a sampled $i$-th trajectories and the size of the vehicle, we evaluate the collision risk for each one. If some part of the vehicle (with an added security margin) is out from the free-space map at any point of a $i$-th trajectory, the whole trajectory is considered to be at risk of collision, and it associated $\alpha_{i}$ is discarded as possible variable for the control action. Fig. \ref{fig:control_action} shows an example of estimated trajectories where the red ones are labeled as collision-risk.

Once we have the collision-free trajectories (green ones in Fig. \ref{fig:control_action}), we evaluate the error between each $j$-th pose in a $i$-th trajectory $\mathbf{x}_{ij}$ and each $k$-th waypoint pose $\mathbf{w}_{k} = \mathbf{p}^{c_k}_{nn}$ in the complete local path $\mathbf{P}^l$. We compute error as follows:

$$
e_{ijk} = c^p \left(\left\lVert\mathbf{x}^p_{ij} - \mathbf{w}^p_{k}\right\rVert\right) + c^o \left(\mathbf{x}^o_{ij} - \mathbf{w}^o_{k}\right) \eqno{(7)}
$$

And we minimizes it as:

$$
i^* = \min_{\forall i} \left(\sum_j \sum_k e_{ijk}\right)
\eqno{(8)}
$$

Where superscripts $p$ and $o$ mean position and orientation, respectively, and where $c^p$ and $c^o$ are configurable constants to relate the different magnitudes. The result $i^*$ is the index for the control action variable $\alpha_{i}$. The variable $v$ of the control action is computed as follows:

$$
v = v_{max} - \left|\alpha_{i}\right| \frac{v_{max} - v_{min}}{\alpha_{max}} \eqno{(9)}
$$

Where $v_{max}$, $v_{min}$, and $\alpha_{max}$, are configurable parameters that depends on the vehicle's characteristics. In case of no collision-free trajectories, control action variables are $\alpha = 0$ and $v = 0$.

%%%%%%%%%%%%%%%%%%%%%%%%%%%%%%%%%%%%%%%%%%%%%%%%%%%%%%%%%%%%%%%%%%%%%%%%%%%%%%%%

\begin{figure}[t]
\centering
\includegraphics[width=220pt]{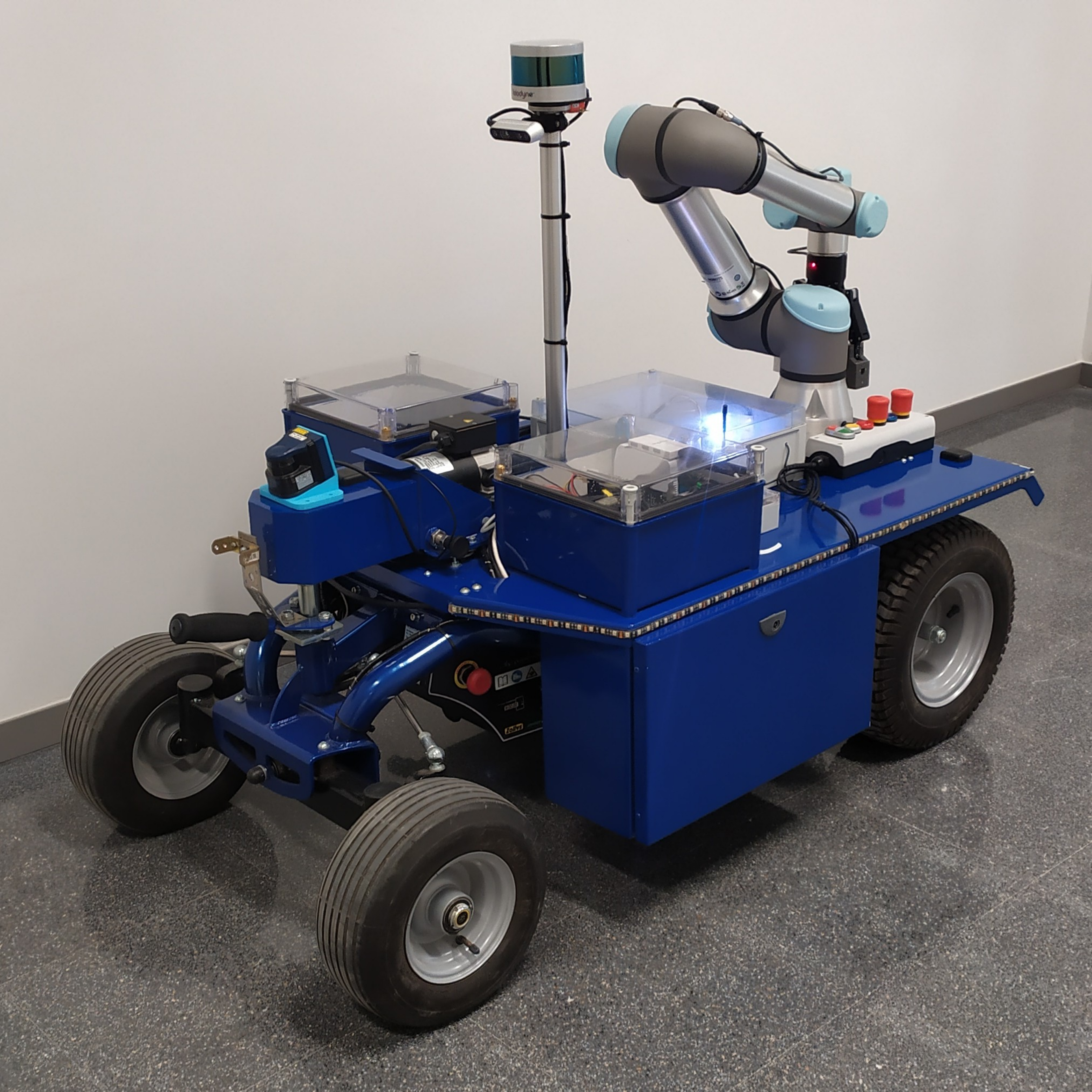}
\caption{Our UGV \textit{BLUE: roBot for Localization in Unstructured Environments} \cite{del2020deeper}.}
\label{fig:blue}
\end{figure}

\section{Evaluation}
\label{sec:experiments}

We evaluated the autonomous navigation system presented in this paper in our research platform \textit{BLUE: roBot for Localization in Unstructured Environments} \cite{del2020deeper} (Fig. \ref{fig:blue}). This robot includes actuators for speed and steering, traction and steering encoders, IMU, GPS Ublox M8P, camera RGBD Intel Realsense D435, and LiDAR 3D Velodyne VLP16. All them integrated into \textit{Robot Operating System} (ROS). The developed software is included in the framework for fast experimental testing presented in \cite{munoz2019framework}. We use a fusion of wheel-encoders, IMU, and GPS as a localization system that provides localization in global coordinates. We use GPS-RTK for ground truth generation, but for the autonomous application, we decided to lighten the context design avoiding dependence on an RTK base station. It is worth noting that besides the mentioned OSM errors, we also have localization errors that could also produce deviations in trajectory. Our NVP module demonstrated robustness against these OSM and localization measurement errors.

We carried out the experiments in the Scientific Park at the University of Alicante. This area contains parking lots and pedestrian walkable areas with trees, benches, and curbs (Fig. \ref{fig:graph}). We chose this scenario for the experiments because it is where the ”pick and place” application commented in the introduction is projected.

In Section \ref{sec:trajectory_ev}, we demonstrate the main advantage of the OSM-based GPP module, which is the possibility of global goal reaching in an extensive area where other approaches such as grid-based ones fail. Also, in Section \ref{sec:trajectory_ev}, we evaluate one of our LPP module (NVP) advantages, which is the navigation in the center of the road, independently of OSM local inaccuracy, compared to another state-of-the-art one \cite{li2021openstreetmap}.  In Section \ref{sec:obstacle_ev}, we evaluate the presented NVP for obstacle avoidance and compare it with the same previous commented method, demonstrating how our system recovers better the center of the road after obstacle avoidance. Finally, in Section \ref{sec:time_ev}, we evaluate the execution time, which is the other main advantage of our NVP, by comparing with \cite{li2021openstreetmap}.

\begin{figure}[t]
\centering
\includegraphics[width=220pt]{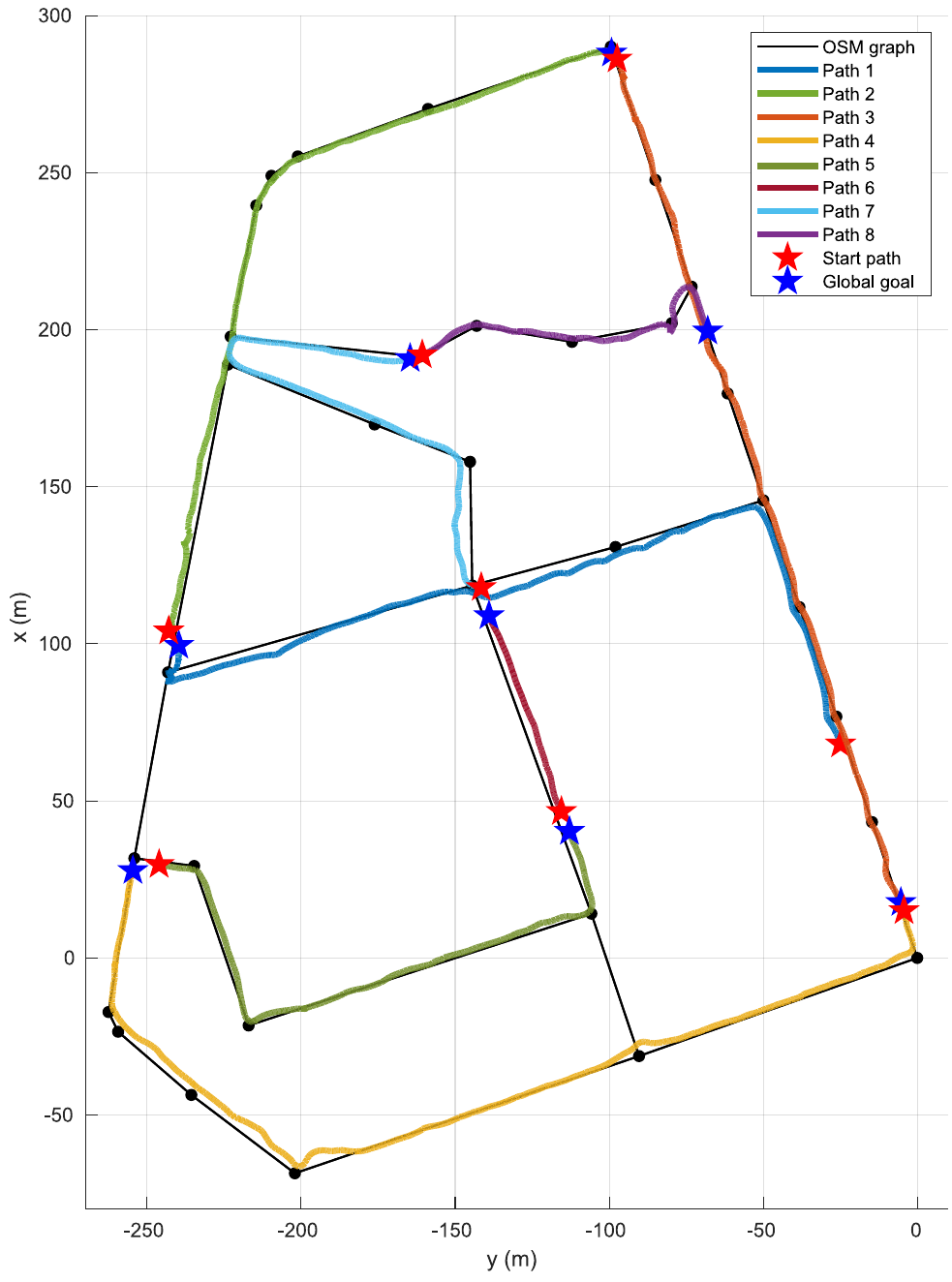}
\caption{Evaluation of global goal-reaching. The red marks indicate the vehicle location when a goal was sent,  and the blue marks indicate the sent goal locations.}
\label{fig:gp_experiments1}
\end{figure}

\begin{figure}[t]
\centering
\includegraphics[width=220pt]{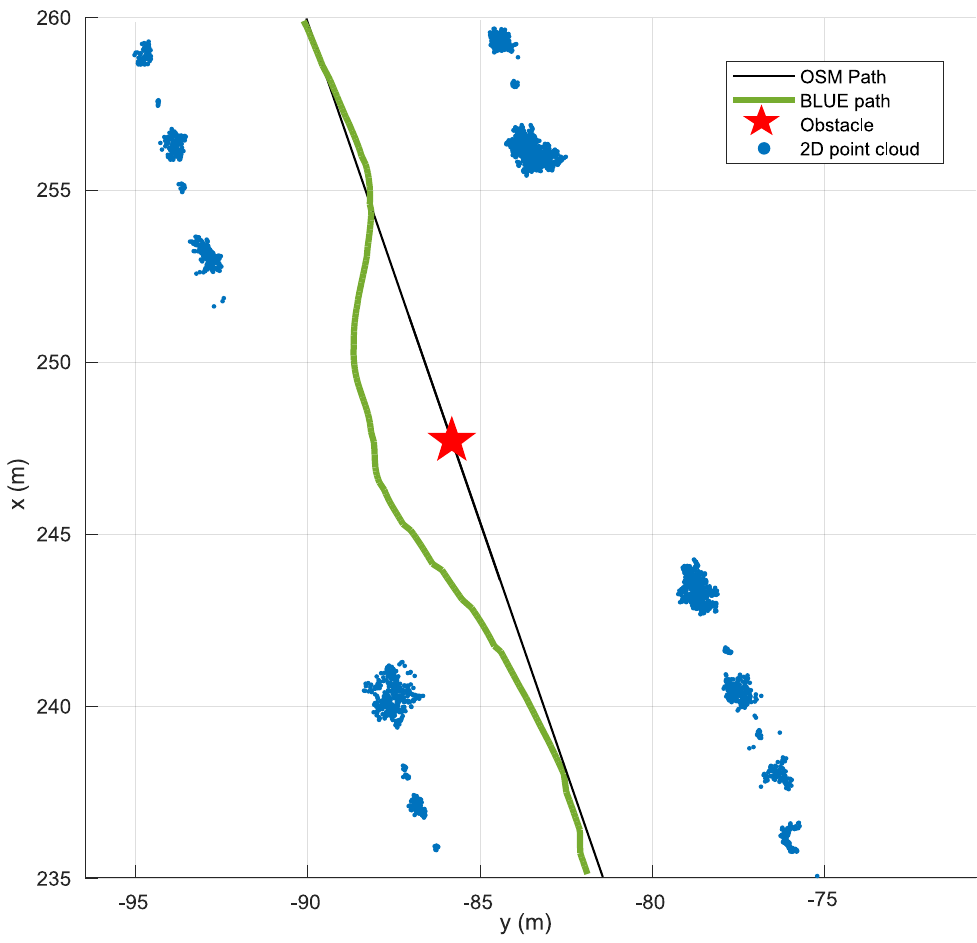}
\caption{Our system behavior in the face of a static obstacle in the center of the road.}
\label{fig:lp_experiments}
\end{figure}

\subsection{Trajectory Evaluation in Global Goal-Reaching}
\label{sec:trajectory_ev}
We test the global goal-reaching for our system using the graph shown in Fig. \ref{fig:graph}. Due to this area being newly constructed, there is no information in OSM concerning trafficable paths. Hence, we created the road network by hand using the JOSM application. Given this graph and the current localization, we send a global goal by hand at a certain point of the network and record the vehicle's autonomous trajectory during the process. Once the goal is reached, we repeat the global goal sending sequentially to cover most of the area of the experiment. In Fig. \ref{fig:gp_experiments1}, we show the results after doing the process described above, where we mark the start of each path and its sent global goal. 

It is worth noting that the road network built is a sparse representation. Then, graph connections have only topological meaning in the context of the GPP layer and don't describe the shape of the road. For this reason, we cannot use the black lines in Fig. \ref{fig:gp_experiments1} as reference for trajectory evaluation. Driving the vehicle through the center of the road following its shape falls on the LPP layer. Then, to evaluate the system quantitatively and compare it with another state-of-the-art one, we drove manually the same paths through the center of the road to use them as ground truth by recording the localization during driving. To obtain enough accuracy, we use (exceptionally for ground truth) a GPS-RTK utilizing a base station on the roof of the main building in the Scientific park area. The accuracy of the RTK system is $0.1m$ for the parts running in RTK-\textit{floating} point mode and $0.02m$ for the places where RTK-\textit{fixed} point is achieved.

Given the ground truth, it was impossible to compare our system with others that use grid maps for GPP because we have problems generating this kind of map in this extensive and highly unstructured environment using maps generators such as Gmapping \cite{grisetti2007improved}. Another challenge using grid maps is to reference it in a global coordinates frame. Such barriers demonstrate the improvement of our GPP module against a vast part of state-of-the-art works \cite{zou2021comparative}. Hence, we compared our approach with another OSM-based autonomous navigation system: \textit{Li et al.} \cite{li2021openstreetmap}. In this work, the authors aim at the problem of OSM error by correcting OSM nodes using a local cost map as a reference.  The LPP layer on \textit{Li et al.} \cite{li2021openstreetmap} uses that cost map constructed from road edge detection from the camera and a classic A$^*$ algorithm for optimization. In Table \ref{tab:gp_experiments}, we show the comparison of error measures against ground truth for both systems. We can see that our method follows better the center of the road, and then its shape, for this environment. The average speed during the experiments was $v_{av} = \SI{1.1}{\frac{\meter}{\sec}}$.

\begin{table}[ht]
\caption{Deviation from the center of the road.}
\label{tab:gp_experiments}
\begin{center}
\begin{tabular}{c c c}
\hline
\textbf{Deviation} & \textbf{NVP-based} & \textbf{Li et al.} \cite{li2021openstreetmap} \\
\hline
Average error (m) & 0.24 & 0.32 \\
Max error (m) & 0.72 & 0.98 \\
\hline
\end{tabular}
\end{center}
\end{table}

\subsection{Obstacle-Avoidance Evaluation}
\label{sec:obstacle_ev}
In the previous section, we evaluated the whole trajectory, which is a task that combines GPP and LPP. However, this trajectory evaluation is for an application in a university campus environment, and, hence, we need to evaluate the behavior of our LPP in the face of unforeseen obstacles.

\begin{figure}[ht!]
\centering
\includegraphics[width=220pt]{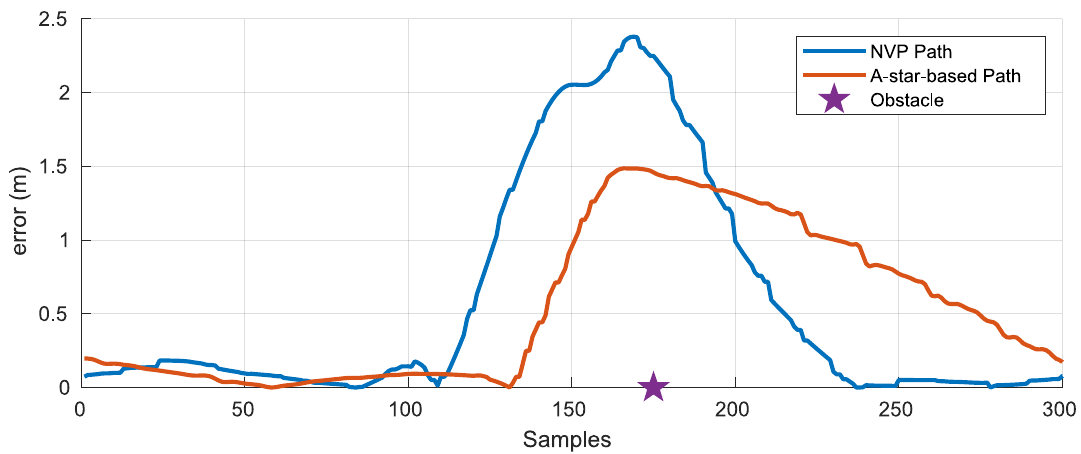}
\caption{Absolute error evolution against the black line in Fig. \ref{fig:lp_experiments} of our system comparing with the error evolution in the same obstacle avoidance for \cite{li2021openstreetmap}.}
\label{fig:lp_error1}
\end{figure}

In Fig. \ref{fig:lp_experiments}, we show our system behavior in the face of a static obstacle in the center of the road. In this case, the black line can serve as a reference, due to, in this case, the road is straight. In order to test how the system recovers the center of the road after obstacle avoidance, in Fig. \ref{fig:lp_error1}, we show the absolute error evolution against the black line in Fig. \ref{fig:lp_experiments} and compare it with the error evolution in the same obstacle avoidance for \cite{li2021openstreetmap}. We can see that our NVP method can recover the center of the road in less time and distance than the tested \cite{li2021openstreetmap} method. Concretely, our method recovers the road's center $83$ samples before the compared one. Given sample time $T = \SI{0.1}{\sec}$, we can assure that our method recovers the center $\SI{8.3}{\sec}$ earlier than the compared one, which means $\SI{9.12}{\meter}$ behind in terms of distance.

\begin{figure}[ht!]
\centering
\includegraphics[width=220pt]{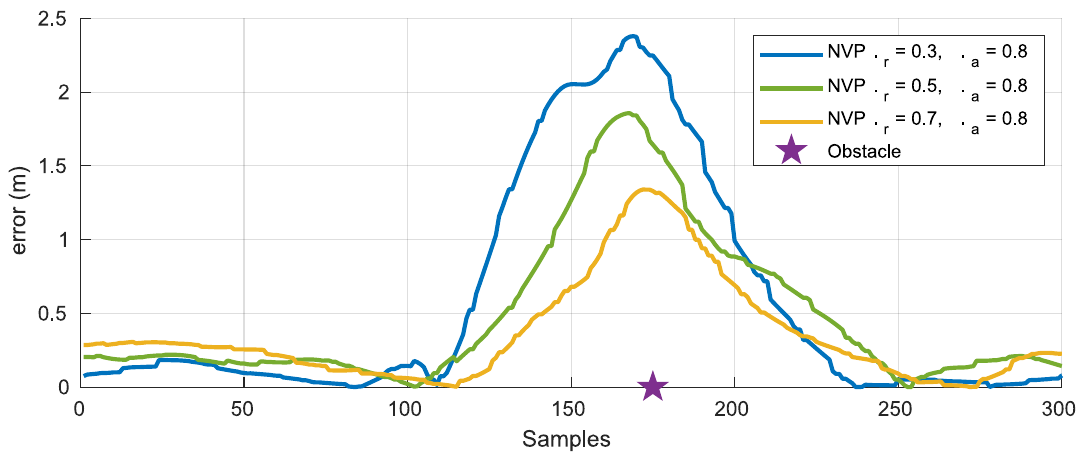}
\caption{Example of different error evolution for different $\gamma_r$ parameter configurations in our NVP potential cost map (4).}
\label{fig:lp_lp_error2}
\end{figure}

How our system avoids obstacles depends on the parameter $\gamma_r$ described in (4), due to it defining the decay of our potential cost map (Fig. \ref{fig:maps} c)). When $\gamma_r > 0$, while $\gamma_r$ increases, the decay in the cost map increases and the trajectory could pass closer to the obstacles. In Fig. \ref{fig:lp_lp_error2}, we show an example of this behavior for three different gamma configurations in the same scenario as the previous experiment. The parameter $\gamma_a$ models the behavior of obstacle avoidance only in the area close to the local goal. For this reason, we configure it constant for the example.

\begin{figure*}[t]
\centering
\includegraphics[width=450pt]{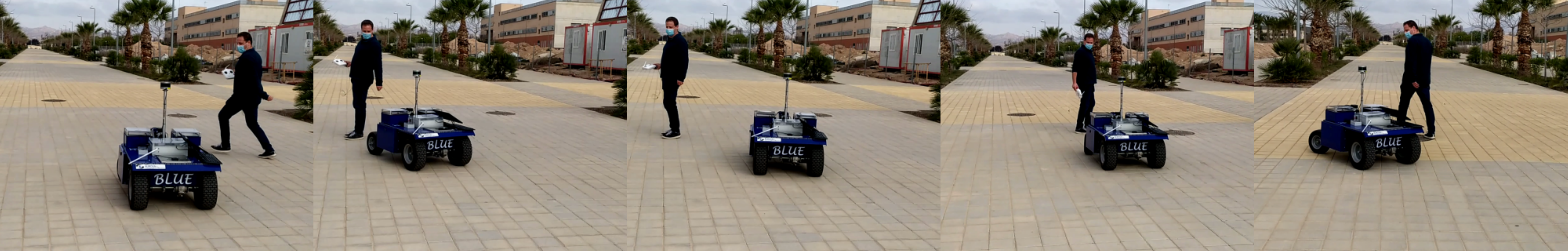}
\caption{\textbf{Obstacle avoidance experiments}: Example of pedestrian avoidance in an image sequence. A pedestrian crosses the vehicle trajectory twice (images 2 and 4 from left to right). In both cases, the vehicle avoids this dynamic obstacle by doing rear maneuvering.}
\label{fig:lp_avoid_1}
\end{figure*}

We also evaluate obstacle avoidance for dynamic obstacles such as pedestrians and vehicles qualitatively. Fig. \ref{fig:lp_avoid_1} shows an image sequence of repeated pedestrian avoidance, where the time evolution is from left to right. We enumerate these images from 1 to 5. In image 1, we can see a pedestrian running to cross the vehicle's trajectory. The vehicle rectifies his local path, but the pedestrian stops just in the front of the vehicle (image 2). In the transition from image 2 to image 3, the vehicle maneuvers, first in the rear direction and after in the front direction, to avoid the pedestrian, but then the pedestrian stops in front of the vehicle again (image 4). Finally, in image 5, the vehicle avoids the pedestrian and continues the travel to reach the goal.

\begin{figure*}[t]
\centering
\includegraphics[width=450pt]{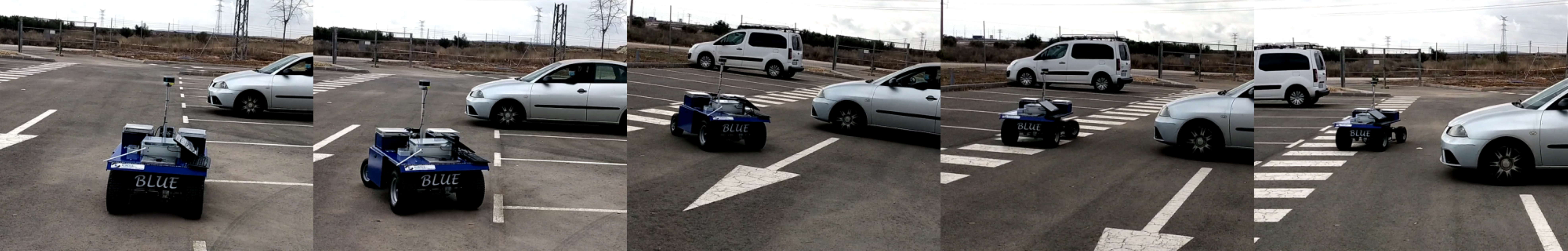}
\caption{\textbf{Obstacle avoidance experiments}: Example of car avoidance in an image sequence. The vehicle circulates in a straight-line way and modifies its trajectory when a car crosses its path.}
\label{fig:lp_avoid_2}
\end{figure*}

In Fig. \ref{fig:lp_avoid_2}, we show another example of obstacle avoidance: a car in a parking lot. In this case, the vehicle follows a straight trajectory, but a car drive to cross this trajectory. However, due to the NVP local path planning module, the vehicle recalculates the path to avoid this dynamic obstacle, as shown in the sequence of Fig. \ref{fig:lp_avoid_2}.

\subsection{Multiple Obstacle-Avoidance Evaluation}
\begin{figure*}[t]
\centering
\includegraphics[width=450pt]{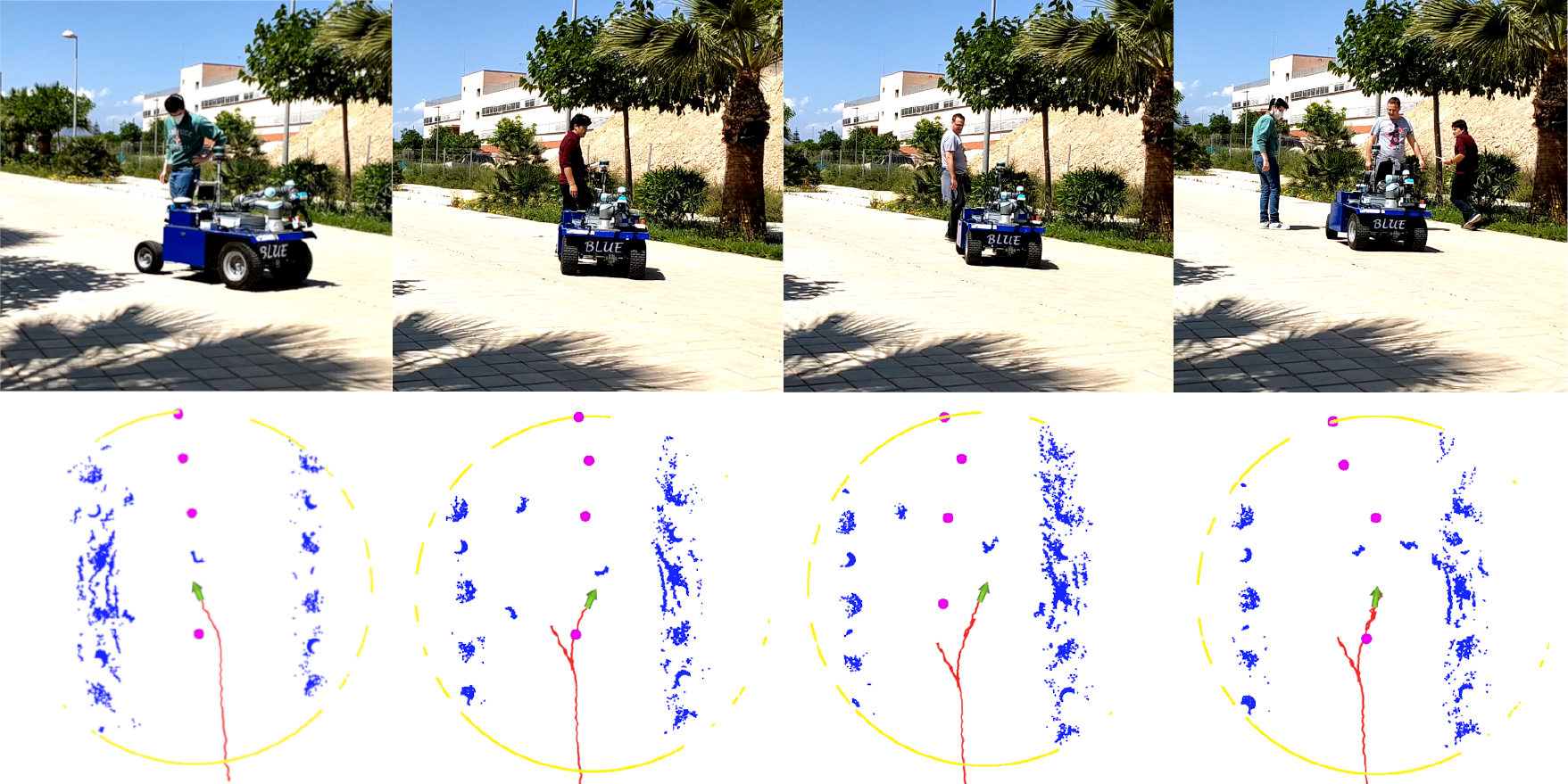}
\includegraphics[width=450pt]{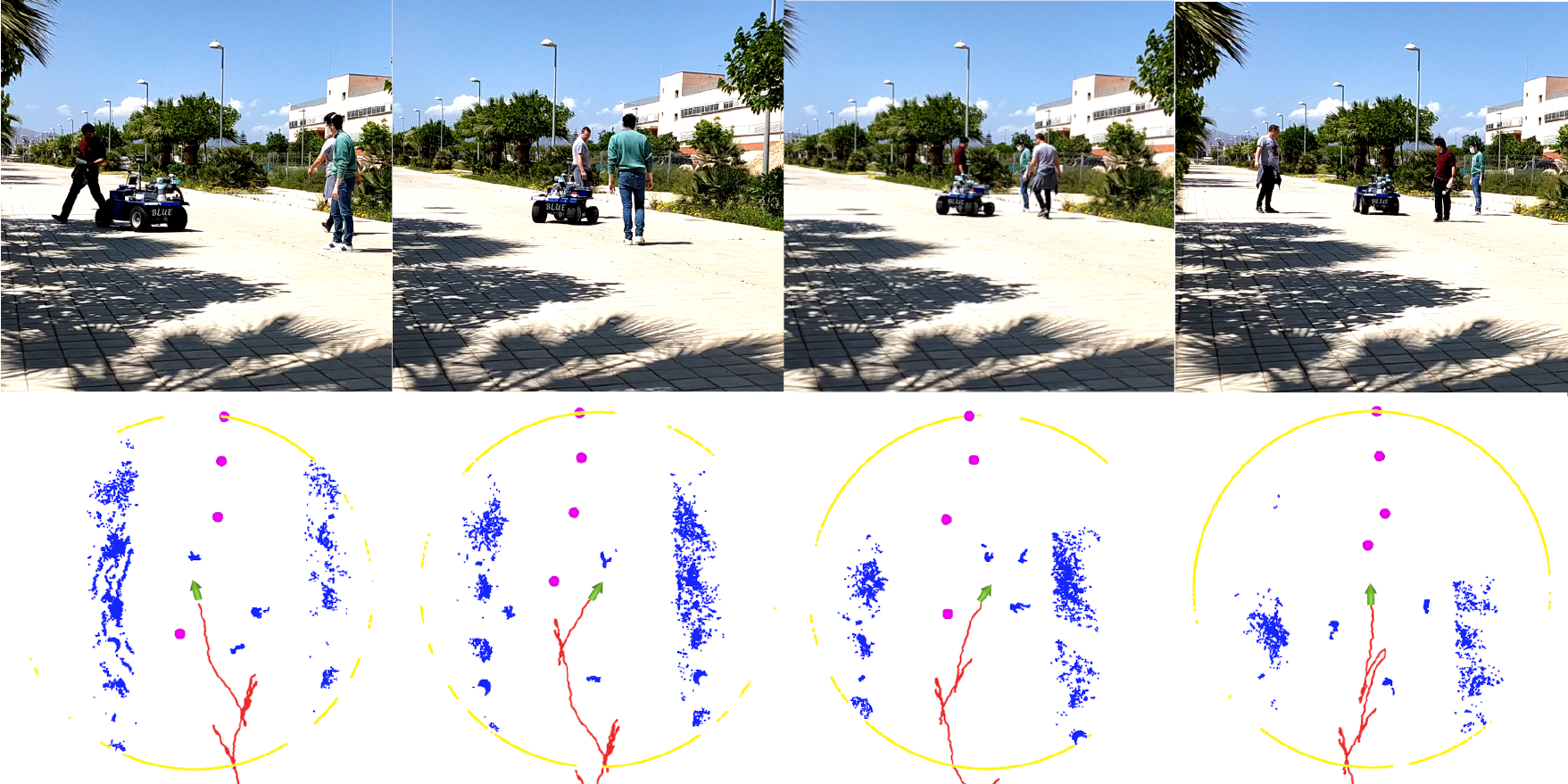}
\caption{\textbf{Multiple obstacle avoidance experiments}: We show the photo sequence of the multiple obstacle-avoidance performed in the upper and third rows. While in the second and bottom rows, we show the corresponding representation of the environment, where the blue points indicate the obstacles, magenta points represent the local path for $N = 4$, the yellow circle draws the outer ring of NVP, and the red line represents the vehicle's trajectory.}
\label{fig:lp_avoid_ped_1}
\end{figure*}

We performed previous experiments with only one obstacle in the environment. However, it can be interesting to evaluate how multiple obstacles affect the NVP. For it, we performed an experiment where three pedestrians crossed the vehicle's trajectory repetitively, simulating even more pedestrians in the environment. For more resolution, we show the results for that experiment divided into different rows in Fig. \ref{fig:lp_avoid_ped_1}. The photo sequence of the  multiple obstacle-avoidance performed is shown in the upper and third rows. While the second and bottom rows shows the corresponding top-view representation of the environment, where the blue points indicate the obstacles, magenta points represent the local path for $N = 4$, the yellow circle draws the outer ring of NVP, and the red line represents the vehicle's trajectory. 

Fig. \ref{fig:lp_avoid_ped_1} shows that multiple obstacle scenarios don't cause problems to our NVP local path planning, and all obstacles were avoided smoothly. We represent the real-time path generation as magenta points for $N = 4$. It is worth noting that when there is an obstacle in the front, the nearest point in the local path is in the  rear part of the vehicle, allowing the rear maneuverings to avoid it.

\begin{table}[ht]
\caption{Execution Time Comparison.}
\label{tab:time_consuming}
\begin{center}
\begin{tabular}{c c c c}
\hline
\textbf{Time} & \textbf{NVP4} & \textbf{NVP8} & \textbf{Li et al.} \cite{li2021openstreetmap} \\
\hline
Average s. time (ms) & 19.8 & 32.8 & 92.7 \\
Max s. time (ms) & 24.7 & 38.5 & 122.8 \\
Total time (s) & 36 & 38 & 53 \\
\hline
\end{tabular}
\end{center}
\end{table}

\subsection{Execution Time Evaluation}
\label{sec:time_ev}

The other advantage of our NVP path planning method is that it is speedy compared to other optimization approaches. To compare the time efficiency, we measured the sample time and total time\footnote{Sample time refers to the execution time of NVP. In contrast, total time refers to the time spent to perform the complete experiment.} of the NVP experiment shown in Fig. \ref{fig:lp_avoid_1}, where the number of circles (Section \ref{sec:naive_path}) is configured $N = 4$ (NVP4). Also, we repeat this process by configuring $N = 8$ (NVP8). Finally, we run the same experiment for the \textit{Li et al.} \cite{li2021openstreetmap} approach. We performed the time measurement for this experiment by running the algorithms as C++ compiled codes on an i7-7700HQ CPU with 16 GB of RAM. We can see in Table \ref{tab:time_consuming} that our approach can complete the scenario shown in Fig. \ref{fig:lp_avoid_1} faster than the \textit{Li et al.} \cite{li2021openstreetmap} one. 

 The average sample time defines the frequency of the method and hence, the maximum velocity of operation. We cannot evaluate the upper-speed limits of the method because our experimental vehicle's maximum velocity is $\SI{1.3}{\frac{\meter}{\sec}}$. However, although the sensor frequency limits the upper-speed de-facto, if we suppose a high-frequency sensor (greater than NVP frequency), we can estimate it analytically as follows. The average sample time of our NVP is $\Delta t = \SI{0.0198}{\sec}$ for $N = 4$. Then, if we consider $d$ as a security margin to avoid an obstacle, we can calculate the approximate upper-speed limit as $v_{up} = d / \Delta t$.

%%%%%%%%%%%%%%%%%%%%%%%%%%%%%%%%%%%%%%%%%%%%%%%%%%%%%%%%%%%%%%%%%%%%%%%%%%%%%%%%
\section{Conclusions And Future Works}
\label{sec:conclusions}

In this paper, we have presented a complete OSM-based autonomous navigation pipeline for Unmanned Ground Vehicles (UGV) in unstructured outdoor environments. As a topological representation, we use road networks from OSM for global path planning. That demonstrates several advantages, such as global consistency and an easy map setup of autonomous navigation applications. At the local path planning level, we presented the novel Naive-Valley-Path method. We demonstrate how this method achieves navigation at the center of the trafficable areas, always following the shape of the road, avoiding the common deviation problems in OSM-based applications. Additionally, given its time efficiency, we show how our NVP method achieves fast and robust obstacle avoidance even in dynamic cases, such as cars and pedestrians, and how it recovers the center of the road after avoidance.

As future works, we plan to use an online interface with JOSM to make a dynamic graph for the GPP. In this way, we could navigate in an exploration mode into completely unknown areas. Also, we plan to research localization using OSM information, and use more sophisticated perception techniques, such a Convolutional Neural Networks (CNN), to classify obstacles and landmarks in the environment.

%%%%%%%%%%%%%%%%%%%%%%%%%%%%%%%%%%%%%%%%%%%%%%%%%%%%%%%%%%%%%%%%%%%%%%%%%%%%%%%%
\bibliographystyle{IEEEtran} 
\bibliography{references.bib}

%%%%%%%%%%%%%%%%%%%%%%%%%%%%%%%%%%%%%%%%%%%%%%%%%%%%%%%%%%%%%%%%%%%%%%%%%%%%%%%%
\begin{IEEEbiography}[{\includegraphics[width=1in,height=1.25in,clip,keepaspectratio]{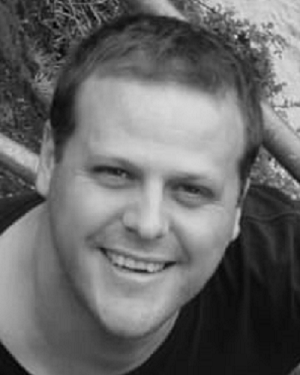}}]{Miguel Ángel Muñoz-Bañón}
received the B.S. degree in telecommunications engineering from the University of Alicante in 2016 and the M.S. degree in artificial intelligence with UNED in 2017. He has worked on different knowledge-transfer projects. He was a Research Technician in differents public projects at Signals, Systems, and telecommunications Group (SST) and Automation, Robotics, and Computer Vision Group (AUROVA) at the University of Alicante. He is currently pursuing a Ph.D. degree in AUROVA at the University of Alicante funding by the Regional Valencian Community Government and the European Regional Development Fund (ERDF) through the grant ACIF/2019/088. His research interests include graph-based simultaneous localization and mapping (Graph-SLAM), geo-referencing using aerial imagery, and machine learning techniques for environment perception.
\end{IEEEbiography}

\begin{IEEEbiography}[{\includegraphics[width=1in,height=1.25in,clip,keepaspectratio]{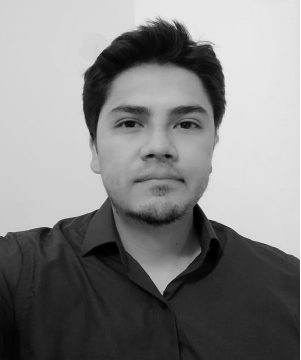}}]{Edison P. Velasco Sánchez}
received the degree in Electronic Engineering and Instrumentation from the University of the Armed Forces ESPE (Ecuador) in 2015 and a Master's Degree in Automation and Robotics from the University of Alicante (Spain) in 2018. He was a research technician in the ARSI research group at the ESPE University and in the Automation, Robotics and Computer Vision Group (AUROVA) of the University of Alicante. He is currently pursuing a Ph.D. degree in AUROVA at the University of Alicante funding by the Regional Valencian Community Government and the Ministry of Science, Innovation and Universities  through the grant PRE2019-088069. His research interests include navigation and autonomous localization in UGVs with cameras and LIDAR sensors.
\end{IEEEbiography}

\begin{IEEEbiography}[{\includegraphics[width=1in,height=1.25in,clip,keepaspectratio]{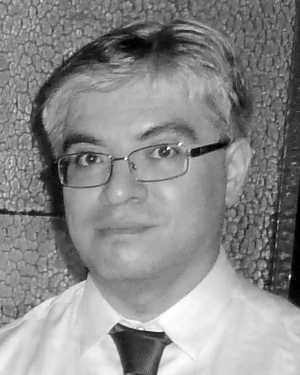}}]{Francisco A. Candelas}
received the Computer Science Engineer and the Ph.D. degrees in the University of Alicante (Spain), in 1996 and 2001 respectively. He is Associate Professor in the University of Alicante since 2003, where he teaches currently courses about Automation and Robotics Sensors in the Degree in Robotic Engineering. Previously, he was in tenure track from 1999 to 2003. Dr. Candelas also researches in the Automation, Robotics and Computer Vision Group (AUROVA) of the University of Alicante since 1998, and he has involved in several research projects and networks supported by the Spanish Government, as well as development projects in collaboration with regional industry. His main research topics are autonomous robots, robot development, and virtual/remote laboratories for teaching.
\end{IEEEbiography}

\begin{IEEEbiography}[{\includegraphics[width=1in,height=1.25in,clip,keepaspectratio]{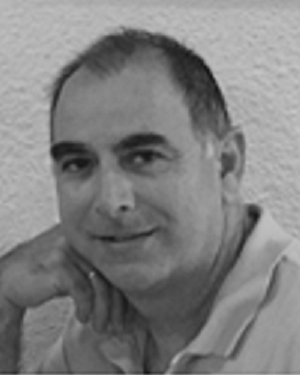}}]{Fernando Torres}
was born in Granada, where he attended primary and high school. He moved to Madrid to undertake a degree in Industrial Engineering at the Polytechnic University of Madrid, where he also carried out his PhD thesis. The last year of his PhD thesis he became a full-time lecturer and researcher at the University of Alicante, and he has worked there ever since. He directs the research group ``Automatics, Robotics and Computer Vision'' founded in 1996 at the University of Alicante. He is a member of TC 5.1 and TC 9.4 of the IFAC, a Senior Member of the IEEE and a member of CEA. Since july 2018 he is coordinator of the area of Electrical, Electronic and Automatic (IEA) of the Spanish Agency of Statal Research (AEI). His research interests include automation and robotics (intelligent robotic manipulation, visual control of robots, robot perception systems, field mobile robots, advanced automation for industry 4.0, artificial vision engineering), and e-learning. Currently, his research focuses on automation, robotics, and e-learning. In these lines, it currently has more than fifty publications in JCR-ISI journals and more than a hundred papers in international congresses. He was Leader Research in several research projects and he has supervised several PhD in these lines of research.
\end{IEEEbiography}

\vfill

\end{document}